\documentclass{article}

\usepackage{arxiv}
\usepackage{xcolor} 
\usepackage{colortbl} 
\usepackage{subcaption}
\usepackage{algorithm}
\usepackage{algorithmic}
\usepackage[utf8]{inputenc} 
\usepackage[T1]{fontenc}    
\usepackage{url}            
\usepackage{booktabs}       
\usepackage{amsfonts}       
\usepackage{nicefrac}       
\usepackage{microtype}      
\usepackage{amsmath} 
\usepackage{lipsum}         
\usepackage{graphicx}
\usepackage{doi}
\usepackage{enumitem}
\title{Optimizing Online Advertising with Multi-Armed Bandits: Mitigating the Cold Start Problem under Auction Dynamics}

\newif\ifuniqueAffiliation
\uniqueAffiliationtrue

\ifuniqueAffiliation 
\author{\hspace{1mm}Anastasiia Soboleva \\
	Avito \\ MSU Institute for Artificial Intelligence 
        \\ Moscow Institute of Physics and Technology\\
	\texttt{soboleva.an@phystech.edu} \\
	\And
	\hspace{1mm} Pudovikov Andrey\\
	Avito\\
        MSU Institute for Artificial Intelligence\\
	\texttt{a.pudovikov@iai.msu.ru} \\
	\AND
        Roman Snetkov\\
	Avito\\
	  \texttt{rvsnetkov@avito.ru} 
        \And
        Alina Babenko\\
	Avito\\
	\texttt{avbabenko@avito.ru} 
        \And
	  \hspace{1mm}Egor Samosvat \\
	  Avito\\
	  \texttt{easamosvat@avito.ru} 
        \AND
	  Yuriy Dorn\\
	MSU Institute for Artificial Intelligence \\ 
        Moscow Institute of Physics and Technology \\
	  \texttt{dornyv@my.msu.ru} \\
    }
\else
\usepackage{authblk}

\newbox{\orcid}\sbox{\orcid}{
\author[1]{%
	\href{https://orcid.org/0000-0000-0000-0000}{\usebox{\orcid}\hspace{1mm}David S.~Hippocampus\thanks{\texttt{hippo@cs.cranberry-lemon.edu}}}%
}
\author[1,2]{%
	\href{https://orcid.org/0000-0000-0000-0000}{\usebox{\orcid}\hspace{1mm}Elias D.~Striatum\thanks{\texttt{stariate@ee.mount-sheikh.edu}}}%
}
\affil[1]{Department of Computer Science, Cranberry-Lemon University, Pittsburgh, PA 15213}
\affil[2]{Department of Electrical Engineering, Mount-Sheikh University, Santa Narimana, Levand}
\fi

\renewcommand{\shorttitle}{\textit{arXiv} Template}

\hypersetup{
pdftitle={Optimizing Online Advertising with Multi-Armed Bandits: Mitigating the Cold Start Problem under Auction Dynamics.},
pdfauthor={Anastasiia Soboleva, Alexander Ledovsky},
}

\begin{document}
\maketitle
\begin{abstract}
Online advertising platforms often face a common challenge: the cold start problem. Insufficient behavioral data (clicks) makes accurate click-through rate (CTR) forecasting of new ads challenging. CTR for "old" items can also be significantly underestimated due to their early performance influencing their long-term behavior on the platform.

The cold start problem has far-reaching implications for businesses, including missed long-term revenue opportunities. To mitigate this issue, we developed a UCB-like algorithm under multi-armed bandit (MAB) setting for positional-based model (PBM), specifically tailored to auction pay-per-click systems.

Our proposed algorithm successfully combines theory and practice: we obtain theoretical upper estimates of budget regret, and conduct a series of experiments on synthetic and real-world data that confirm the applicability of the method on the real platform. 

In addition to increasing the platform's long-term profitability, we also propose a mechanism for maintaining short-term profits through controlled exploration and exploitation of items.
\end{abstract}

\section{Introduction}

Predicting click-through rates (CTR) is a critical task in e-commerce and digital marketing, particularly in online advertising systems, where it serves as a key metric for determining ad rankings in online auctions. By accurately predicting CTR, ad platforms can optimize the business metrics, improve return on investment (ROI) of their advertisers, and ultimately increase revenue.

The prevailing approach to evaluating CTR relies on the application of deep learning techniques \cite{cheng2016wide, wang2017deep, zhou2018deep, guo2017deepfm}, leveraging historical user interactions with items, such as clicks, which serve as critical training signals for CTR prediction. However, a critical limitation arises when newly introduced and dormant products lack sufficient behavioral data, rendering them susceptible to being underestimated and relegated to the long tail of search results or even entirely omitted from the query output. This leads to a cause-and-effect dilemma: the lack of behavioral data leads to poor ranking, which in turn leads to new products being less likely to receive behavioral data. The phenomenon is referred to as the item's "cold start problem" \cite{ye2023cold, zhu2019addressing, han2022addressing, lika2014facing}, a pervasive issue that has significant implications for businesses, including unrealized opportunities for increased platform revenue generation.

On the other hand, merely increasing traffic to new ads during cold start is not an effective solution, as it may come at the expense of short-term and long-term profitability metrics and key performance indicators (KPIs), including customer satisfaction and seller retention. This situation necessitates a trade-off between two goals: maximizing immediate returns through the selection of established ads (exploitation) and achieving long-term growth through the introduction of cold ads (exploration).

The exploration-exploitation dilemma fits perfectly into the sequential decision-making framework, and more specifically, the multi-armed bandit (MAB) setting \cite{slivkins2019introduction}. The MAB approach has been extensively explored in literature, yielding a suite of efficacious algorithms, including the Upper Confidence Bound (UCB) \cite{auer2002finite} and Thompson's Sampling (TS) \cite{gupta2011thompson}. These methods have been generalized and applied to diverse domains, such as online advertising \cite{feng2023improved}, 
recommendation systems \cite{mary2015bandits, silva2022multi} and social networks \cite{sankararaman2019social}.

In the context of pay-per-click auction systems on online advertising platforms, ads (arms) are viewed as entities competing for opportunities (positions) in auctions. Each arm declares its price per click. However, the reward ($price \cdot CTR$) associated with each arm remains uncertain due to the unknown CTR.

Previous research on cold start in online advertising has focused primarily on improving CTR prediction accuracy and/or learning rates. However, existing solutions overlook the pay-per-click (PPC) auction systems that dominate online platforms. We consider the economic aspects of cold start and aim to explicitly model cold start as an important lever for improving long-term business income. Our objective is to develop an online algorithm that effectively allocates ad spaces for advertisers in cold-start environments, maximizing their return on investment while addressing the unique challenges of PPC advertising platforms.

Additionally, theoretical results for the MAB setting under positional-based model (PBM) \cite{craswell2008experimental} with different prices distributions among sellers are limited in the previous works, which motivates us to fill this gap in our work.

To address the context of online advertising, we developed a UCB-like algorithm under MAB setting for PBM, specifically tailored to auction pay-per-click systems. The key results of this paper are summarized as follows:
\begin{itemize}

\item We focus on the economic aspect of the cold-start problem, whereas previous studies have primarily concentrated on enhancing the accuracy of click-through rate (CTR) forecasting and maximizing the number of clicks.
\item  We develop a UCB-like algorithm under MAB setting for positional-based model, specifically tailored to auction pay-per-click systems.
\item  We conduct a series of experiments on synthetic and real industrial data that confirm the applicability of the method.
\item  We obtain theoretical performance guarantees, specifically an upper bound on budget regret.
\item  We propose an approach to optimize short-term profitability by controlling exploration strategies, leveraging  CTR estimates from our existing production model.
\end{itemize}

Our work is closely related to several recent studies \cite{zhou2023bandit,lagree2016multiple, feng2023improved}, which have explored various aspects of this problem. Notably, \cite{feng2023improved} focused on CTR estimation for single-slot pay-per-click auctions, while we extend their approach to consider multiple slots. Although other research \cite{zhou2023bandit,lagree2016multiple} have also addressed scenarios with multiple slots and employed a position-based model (PBM), they overlooked the complexities inherent in pay-per-click auctions. 

To address these gaps, we propose \textbf{AuctionUCB-PBM}, a UCB-like algorithm tailored to multi-armed bandit settings under PBM, specifically designed for auction pay-per-click systems. Our study bridges the gap between theoretical learning guarantees and practical advertising cold-start challenges, offering a straightforward and easily implementable solution for real-world online advertising platforms.

\section{Related Work}
Our paper is primarily related to four streams of literature: CTR prediction, item's cold start, multi armed bandit (MAB), position based model (PBM).

\textit{CTR prediction.} Click-through rate (CTR) prediction is one of the most central tasks in online advertising systems. Recent deep learning-based models that exploit feature embedding and high-order data have shown dramatic successes in CTR prediction. These approaches include Deep Interest Network (DIN)\cite{zhou2018deep}, Deep\&Cross Network  (DCN) \cite{wang2017deep}, Wide\&Deep \cite{cheng2016wide}, and DeepFM \cite{guo2017deepfm}. These models have shown dramatic successes in CTR prediction by automatically learning latent feature representations and complex interactions between features in different ways. However, these models work poorly on cold-start ads with new IDs, whose embeddings are not well learned yet.

\textit{Cold start.} The majority of existing solutions to the item cold-start problem rely on a content-based approach, leveraging the characteristics of new items to discover analogous user preferences and recommending these items accordingly. Various techniques \cite{agarwal2009regression, koren2008factorization} have been developed to adapt matrix factorization (MF) methods for cold-start scenarios by incorporating item-specific attributes, such as item's descriptions and contents, into the model. These extensions generate vector embeddings that can be compared with user representations from a lookup table to provide personalized recommendations. As a complementary approach to collaborative filtering, content-based filtering \cite{lops2011content} offers a distinct strategy for recommending new items. By analyzing the attributes and characteristics of each item and comparing them with those of items previously interacted with by the user, this method can identify relevant similarities and provide personalized recommendations. The integration of collaborative filtering and content-based filtering techniques has given rise to hybrid systems \cite{wei2016collaborative}, which are designed to enhance the accuracy and reliability of click-through rate (CTR) predictions. By combining the strengths of both methods, these hybrid systems can provide more accurate and robust predictions. This synergy also enables them to effectively address cold-start challenge. Furthermore, researchers  \cite{vartak2017meta, zheng2021cold} have employed meta-learning approaches \cite{vilalta2002perspective} that enable the exploration of prior knowledge across various tasks and facilitate the development of reasonable initial parameters for cold items. The paper \cite{panda2022approaches} undertakes a comprehensive systematic literature review of research efforts between 2010 and 2021 to address click-through rate prediction and cold start problems. The review synthesizes a diverse range of approach-driven strategies, including deep learning, matrix factorization, hybrid methods, and innovative techniques in collaborative filtering and content-based algorithms, providing a better understanding of the state-of-the-art solutions.

\textit{MAB.} In contrast to existing approaches, our method operates within a multi-armed bandit (MAB) framework \cite{auer2002finite, slivkins2019introduction}, which enables incremental feedback without requiring additional data or using any neural network architecture. The success of MAB algorithms relies heavily on their ability to balance exploration and exploitation in their action selection policies. If an agent prioritizes exploration, it may overlook valuable insights by randomly choosing new actions without considering the knowledge gained from previous steps. Conversely, if an agent focuses solely on exploitation, it will favor short-term rewards over long-term benefits, neglecting optimal solutions.

To address this critical trade-off, researchers have developed a range of MAB algorithms with unique properties and strengths. Some of the most well-established methods include $\epsilon$-Greedy \cite{auer2002finite}, which strikes a balance between exploration and exploitation by selecting actions randomly with probability $\epsilon$. Upper Confidence Bounds (UCB) \cite{auer2002using} uses confidence bounds to determine which action to take, balancing exploration and exploitation based on the uncertainty associated with each option. Thompson Sampling (TS) \cite{chapelle2011empirical} is a probabilistic approach that takes into account both prior knowledge and observed data to make informed decisions.

\textit{PBM} Research in web search relies on understanding user behavior. For example, \cite{joachims2017accurately} demonstrated that users consistently favor top-ranked items over lower-ranked ones on a search engine results page (SERP), with a clear preference for the first item listed. The study \cite{chuklin2022click} presents different basic click models that capture different assumptions about searchers' interaction behavior on a web search engine result page, including the PBM (positional-based model). The PBM posits that users click an item only if they have viewed the item and drawn by its relevance. This model is also grounded in the intuitive notion that the probability of a user examining an item depends heavily on its position on a SERP, typically decreasing with rank.

Our work is closely related to several recent studies \cite{zhou2023bandit, lagree2016multiple, feng2023improved} on online advertising auctions. Specifically,  \cite{feng2023improved} addresses CTR estimation for single-slot Vickrey-Clarke-Groves auctions, while we extend this approach to consider multiple slots. Although \cite{lagree2016multiple, zhou2023bandit} also consider scenarios with multiple slots, they utilize a position-based model(PBM) and neglect the complexities associated with pay-per-click auctions. Our research aims to address these gaps by developing \textbf{AuctionUCB-PBM}, a UCB-like algorithm under MAB setting for PBM, specifically tailored to auction pay-per-click systems. 

\section{Problem Formulation}

\subsection{Preliminaries}
We consider an online advertising platform that employs a multi-slot first-price auction mechanism. We assume a repeated $K$-slot ad auction setting, where there are $T$ rounds and $K$ ads displayed in each round. For simplicity, we consider $T$ to be a fixed and known.

In the standard pay-per-click model, advertisers only incur costs when their ads are clicked by users. To maximize revenue, an advertising system must select a subset of candidate ads to display and rank them optimally. This requires predicting the expected revenue for each ad, which depends on both the price $P_k$ and CTR $\theta_k$ of each item $k\in [K]$. Specifically, each ad $k\in [K]$ is characterized by its expected cost per impression ($eCPI_k$), denoted as $eCPI_k := P_k \cdot \theta_k $. We sort the ads in descending order based on this value to display them accordingly. While the advertisers' prices are known, the click-through rates can vary significantly. Therefore, to run an efficient first-price pay-per-click auction, we need to estimate the CTR for each ad.

The learner's objective is to maximize the total budget reward, which is equivalent to minimizing regret. Regret is defined as the difference between the cumulative reward achieved in the optimal policy and the total reward of the chosen policy. The optimal policy refers to a scenario where the player (platform) knows the expected rewards (perfect knowledge of the true CTRs of all items) in advance and always selects the optimal ads' ranking in each round.

The order in which items are presented has a significant impact on their visibility. Consequently, it is crucial to consider the bias introduced by the visualization mechanism. A natural choice for user modeling is the position-based model (PBM), which posits that each ranking position is associated with a probability of being observed. 

\subsection{MAB Setting}
Although we consider the task of ranking, where the number of items equals the number of slots. For convenience in notation, we denote positions as $l \in [L]$ and items as $k \in [K]$, where $[K] := \{1,2,...,K\}$.
\subsubsection{Setting}
At each round $t \in [T]$, the learner chooses an ordered list of $K$ arms (items), denoted as an action. The set of all possible actions is represented by $\mathcal{A}$ and contains $K!$ unique ordered lists. At time $t$, the selected action will be denoted as $a(t) = (a_1(t),\ldots, a_L(t))$, where each $a_l(t)$ corresponds to the item occupying position $l \in [L]$.

In its most classical formulation \cite{auer2002finite, slivkins2019introduction}, a MAB problem is defined by independent and identically distributed random variables $X_k(t)\sim\mathcal{B}(\theta_k)$ for each arm $k \in [K]$ and time step $t \in [T]$. Here, $\mathcal{B}(\theta)$ denotes the Bernoulli distribution with parameter $\theta$, representing the probability of a click or conversion. Specifically, each arm $k$ is associated with an unknown expected reward $\theta_k\in[0,1]$.

When considering PBM, we enrich this formulation by incorporating known positional discount rates $(\gamma_l)_{l=1}^L$, where each $\gamma_l \in [0,1]$ represents the probability that a user effectively observes an item in position $l$. We assume that these discount rates are strictly decreasing, i.e., $\gamma_1 > \gamma_2 > \ldots > \gamma_L$.

Due to the PPC system, each item $k\in [K]$ is associated with a fixed price $P_k$ for a single click.

\subsubsection{Regret}
At each round $t\in[T]$, the learner selects an action $a(t)\in \mathcal{A}$, which is then presented to the user. Then the learner receives a reward $r_{a(t)}$ that can be characterized by its expected value:
 $$
 \mathbb{E}r_{a(t)} = \sum_{l=1}^L  \gamma_l\cdot P_{a_l(t)}\cdot\theta_{a_l(t)}
 $$ 
The objective is to maximize the cumulative reward over time, which is equivalent to minimizing regret. Regret is defined as the difference between the cumulative reward that would have been achieved if the decision maker had always chosen the optimal action, and the actual cumulative reward obtained.

To simplify notation, we assume that $P_1\theta_1>P_2\theta_2>...>P_K\theta_K$. The fact that the sequences $(P_k\theta_k)_{k = 1,..., K}$ and $(\gamma_l)_{l = 1,...,L}$ are in decreasing order implies that the optimal list is $a^* = (1,...,K)$. Denoting by $R(T) =\sum_{t=1}^T \left (r_{a^*}-r_{a(t)} \right )$ the regret incurred by the learner up to time $T$, one has

$$\mathbb{E}[R(T)] = \mathbb{E}\left[\sum_{t=1}^T\sum_{l=1}^L\gamma_l(P_{a^*_l}\theta_{a^*_l}-P_{a_l(t)}\theta_{a_l(t)})\right] =  $$

$$ = \sum_{a\in \mathcal{A}}\Delta_a\mathbb{E}[\tau_a(T)] \rightarrow \min$$ 
where $\Delta_a$ is the expected gap of action $a$ to optimality: $\Delta_a=\sum_{l=1}^L\gamma_l\left(P_{a^*_l}\theta_{a^*_l} - P_{a_l}\theta_{a_l(t)}\right)$, and, $\tau_a(T) = \sum_{t=1}^T\mathbb{I}\{a(t)=a\}$ is the number of times action $a\in \mathcal{A}$ has been chosen up to time $T$.

\subsubsection{Arm Mean Estimator.}
For further analysis, we define two statistics:
\begin{itemize}

\item $S_k(t)$ is the \textbf{cumulative target actions} statistic, which counts the total number of target actions brought to the learner by arm $k\in[K]$ up to moment $t$: $ S_k(t) = \sum_{s=1}^{t} X_k(s)$

\item $N_k(t)$ is the \textbf{effective impressions} statistic, which counts the total number of effective impressions of item $k\in [K]$ by time $t$:
$N_k(t) = \sum_{l= 1}^{L} \sum_{s=1}^{t-1}\gamma_l \mathbb{I}\{a_l(s) = k\}$, where $\mathbb{I}\{\}$ is the indicator variable of the event.
\end{itemize}

We note that $N_k(t)$ depends on both the ranking history $\{a(t)\}_t$ and the position parameters $(\gamma_l)_{1\leq l\leq L}$. Then, for arm $k\in [K]$, the asymptotically unbiased arm mean estimator can be defined as $$\hat\theta_k(t) = \frac{S_k(t)}{N_k(t)}$$.

\textbf{Proposition 1.} \textit{For arm $k \in [K]$, denote the unbiased empirical arm means by $\hat\theta_k(t) = \frac{S_k(t)}{N_k(t)}$. Then, using Chernoff-Hoeffding Inequality
\cite{chernoff1952measure, hoeffding1994probability}, for any $\delta > 0$ we have}
$$\mathbb{P}\left( \left|\hat \theta_k(t) - \theta_k\right| \geq \sqrt{\frac{\delta \ln t}{N_k(t)}}\right)\leq 2\cdot t^{-2\delta}.$$

\section{UCB Strategy under PPC system} 

The primary objective of UCB strategy is to apply the upper bound as a measure of potential for each arm based on the uncertainty surrounding its quality. In this context, the learner consistently selects the most promising action with the highest upper confidence bound $U_k(t)$. This balances exploration (acquiring new information) and exploitation (optimizing for immediate rewards).

We employ the upper confidence bound formula for each arm $\hat\theta_k(t)$, based on Proposition 1:

$$U_k(t) = \frac{S_k(t)}{N_k(t)} + \sqrt{\frac{\delta \ln t}{N_k(t)}}.$$

The learning agent interacts with the environment for $T$ rounds. At each time step $t \in [T]$, it selects an action $a(t)$ that maximizes its expected utility, based on the upper confidence bounds of all available actions. Specifically, the agent chooses actions in descending order of their corresponding upper confidence values: 
$$P_{a_1(t)}U_{a_1(t)}(t) \geq P_{a_2(t)}U_{a_2(t)}(t) \geq ... \geq P_{a_L(t)}U_{a_L(t)}(t).$$

\subsection{Regret Analysis}
\textbf{Theorem 1.} 
\label{theorem1}
\textit{Let} $\Delta_{min} = \min\limits_{a\in \mathcal{A} : \Delta_a \neq 0} \Delta_a$, $\Delta_{max} = \max\limits_{a\in \mathcal{A}}\Delta_a$,\\  $C(\gamma)=\min\limits_{l=1,..,K}\frac{\left(\sum_{k=1}^K \gamma_k\right)^2}{l}+\left(\sum_{k=1}^l \gamma_k\right)^2$, \textit{and} $P_{max} = \max\limits_{k\in[K]}P_k$. \textit{Using L2R-BudgetUCB with $\delta = 1.5$ the regret is bounded from above by $$\mathbb{E}\left[R(T)\right] \leq \frac{\pi^2}{3}K\Delta_{max} + \frac{ 64 K C(\gamma) P_{max}}{\Delta_{min}} \ln T.$$}
\\\textbf{Proof.} For a detailed and rigorous proof, please refer to the $Appendix~A$. Here, we provide an overview of the key steps involved.

The proof of $Theorem~1$ relies on regret decomposition:  $$R(T) = \sum_{t=1}^T \Delta_{a(t)}\mathbb{I}_{E_t} + \Delta_{a(t)}\mathbb{I}_{\bar E_t},$$  where $E_t = \left \{\exists k \in a(t) : \left|\hat \theta_k(t) - \theta_k\right| \geq \sqrt{\frac{\delta \ln t}{N_k(t)}} \right\}$ associated with the event that $\theta_k$ is outside of the high-probability confidence interval around $\hat\theta_k$.

The first term in our regret decomposition can be bounded from above in expectation by a constant that does not depend on $T$: $\sum_{t=1}^T \Delta_{a(t)}\mathbb{I}_{E_t}\leq \frac{\pi^2}{3}K \Delta_{max}$. 

We evaluate the second term by adapting the calculations presented in \cite{lagree2016multiple, combes2015combinatorial, kveton2015tight} to our model. Specifically, for all times $t \in [T]$ we define the set of arms $S_t$ that are not observed "sufficiently often" up to time $t$ as:
\\$S_t = \{1\leq k \leq K : N_k(t)\leq \frac{16 \delta\ln T P_{max} \left(\sum_{k=1}^K\gamma_k \right)^2}{\Delta^2_{a(t)}}\}$ 
\\and the related events:
\\ $F_t = \left \{0 < \Delta_{a(t)}\leq 2\sum_{k=1}^K\gamma_k P_{a_k(t)}\sqrt{\frac{\delta \ln T}{N_k(t)}} \right \};$
\\ $G_t = \{\left|S_t\right|\geq l\};$
\\ $H_t = \left \{\left|S_t\right|< l\right \} \cap \left \{ \exists k \text{ s.t. } N_k(t) \leq \frac{16 \delta P_{max}\left(\sum_{k=1}^l\gamma_k \right)^2}{\Delta^2_{a(t)}} \ln T\right\}$.
\\In event $H_t$  the constraint on $N_k(t)$ only differs from the first one ($S_t$) by its numerator which is smaller than the previous one, leading to an even stronger constraint. $F_t$ is the event that suboptimal action $a(t)$ is "hard to distinguish" from $a^*$ at time $t$. In the $Appendix~A$ we provide a detailed proof of all the calculations and the following chain of inequalities:
\begin{equation*}
    \begin{split}
        &\sum_{t=1}^T\Delta_{a(t)}\mathbb{I}\{\bar E_t\} \leq \sum_{t=1}^T\Delta_{A(t)}\mathbb{I}\{F_t\} \\
        &\leq \sum_{t=1}^T \Delta_{A(t)}(\mathbb{I}\{G_t\} + \mathbb{I}\{H_t\}) \leq \frac{ 64 K C(\gamma) P_{max}}{\Delta_{min}} \ln T.
    \end{split}
\end{equation*}.\hfill

\subsection{Algorithm.}
\label{section:algorithm}

We will run pay-per-click first-price auctions by utilizing the Upper Confidence Bound (UCB) estimates of the click-through rates (CTRs). We defer the details of this UCB-style online mechanism to $Algorithm~1$. Following a standard regime in UCB algorithms, we initiate a forced exploration phase for each arm at the beginning to obtain a warm start for the main UCB online mechanism.

In our regret analysis, we disregard the regret incurred during the initial exploration phase since it can be easily bounded by a constant.

The main logic of algorithm is following UCB strategy and the proposed theorem \ref{theorem1}. To use this theorem next constraint must be satisfied: $N_k(t) > 0$, which can be achieved by allocating one opportunity for each item. Also, there is no dependence on clicks. It can be implemented by one random ordering (see line \ref{algothm::line1}) and then log results (line \ref{algothm::line2}) of corresponding action. Then during the cycle you should use formula for $U_k(t)$ (line \ref{algothm::line4}) to sort (line \ref{algothm::line5}) your items and then show to user (line \ref{algothm::line6}). In the end, all that is left to do is update the parameters (lines \ref{algothm::line7}, \ref{algothm::line8}). You can see total schema in Algorithm \ref{alg:AuctionUCB-PBM}.

\begin{algorithm}[h] 
\caption{AuctionUCB-PBM}
\label{alg:AuctionUCB-PBM}
    \textbf{Input}: K different ads, K visibilty coefficients sorted in descending order, $T$ - number of steps, $\delta = 1.5$ - parameter for UCB increament\\
    \begin{algorithmic}[1] 
    \STATE \label{algothm::line1} Show all ads in random order with visibility $\gamma_k$, \\
    \STATE \label{algothm::line2} Get target actions $r_k$, set $t=1, S_k(1) = r_k$, $ N_k(1) = \gamma_k$ for $k \in [K]$
    \WHILE{$t < T$}
    \STATE \label{algothm::line4} Get $S_k(t), N_k(t)$ and then calculate $U_k(t)$ for each arm $k \in [K]$
    \STATE \label{algothm::line5} Sort $U_k(t) \cdot P_k$ in descending order, get $\gamma_k$ for all arms
    \STATE \label{algothm::line6} Show ads in such order to system, get rewards (target actions) $r_k$
    \STATE \label{algothm::line7} $t = t + 1$
    \STATE \label{algothm::line8} Log additional visibility $N_k(t) = N_k(t-1) + \gamma_k$ and target actions $S_k(t) = S_k(t-1) + r_k$ for all arms
    
    \ENDWHILE
    \end{algorithmic}
\end{algorithm}

\section{Experiments}
\label{sec:experiments}

The algorithm is evaluated through two distinct experiments. The first section \ref{synthetic_experiment}, demonstrates effectiveness using entirely synthetic data. The second section \ref{real_data_experiment}, evaluates performance with real data from Platform O. Both sections focus on the case where $K = 30$ and utilize identical visibility constants $\gamma_l$ from the real dataset.

A key consideration in applying the Multi-Armed Bandit (MAB) approach is the initialization of each object. For the algorithm, ensuring non-zero cumulative visibility is crucial, regardless of recorded clicks as discussed in section \ref{section:algorithm}. Given the significant variability among objects in real-world contexts and the inability to provide users with random or poorly-performing advertisements, predictions from machine learning-based Click-Through Rate (CTR) models establish the initial opportunity for each item.

Throughout the simulations, the results of the experiments will be evaluated only in terms of their mathematical expectation of $Regret$ and $Regret/t$. To obtain correct results, all experiments were averaged at least 24 times.

\subsection{Synthetic}
\label{sec::synthetic}
This section generates synthetic data to evaluate the algorithm. A total of $K=30$ distinct items are considered at each step, followed by the application of the algorithm outlined in \ref{alg:AuctionUCB-PBM}.
\label{synthetic_experiment}
\begin{figure}[h]
    \centering
        \includegraphics[width=0.5\linewidth]{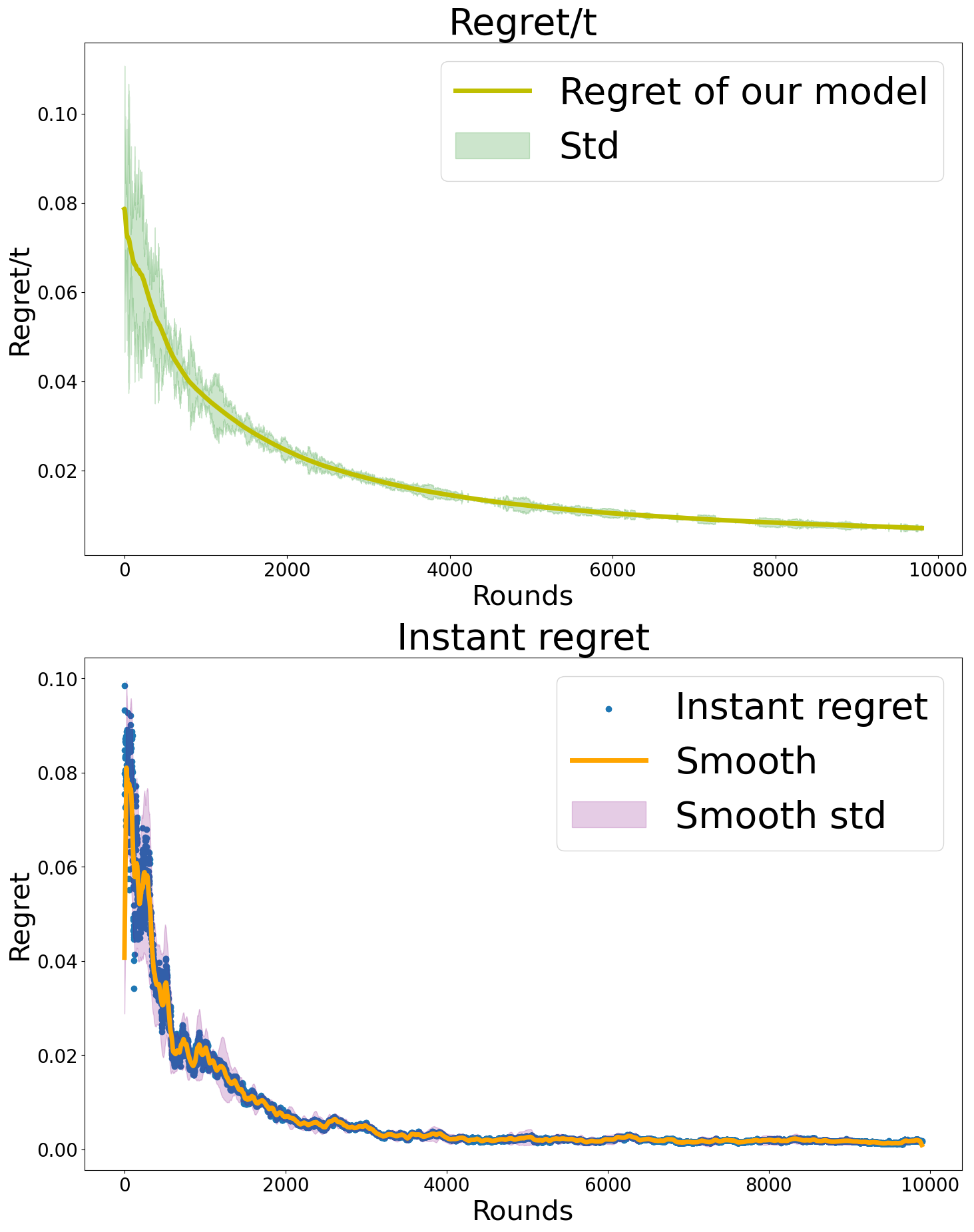}
    \caption{Evaluation of AuctionUCB-PBM on synthetic data, generated as 1(c) combination - fixed $price=1$ and CTR's from real distribution. Upper is average Regret/t of our algorithm and lower is instant regret per round.}
    \label{fig::synthetic1}
\end{figure}

The principal aim is to validate convergence to the optimal solution, estimated using various probability distributions for eCPI's. As eCPI comprises conversions and price, different marginal distributions for these factors are analyzed. However, only the joint distribution of the product of conversions and price is used for determining the optimal order.

A significant challenge lies in accurately learning the true eCPIs for each item, equating to learning conversions. If the eCPI distribution is concentrated around a single point with low variance (particularly in the first two decimal places), more simulations will be required to accurately establish the true order due to the limited number of actual conversions from displayed advertisements.

\begin{figure}[h]
    \centering
        \includegraphics[width=0.5\linewidth]{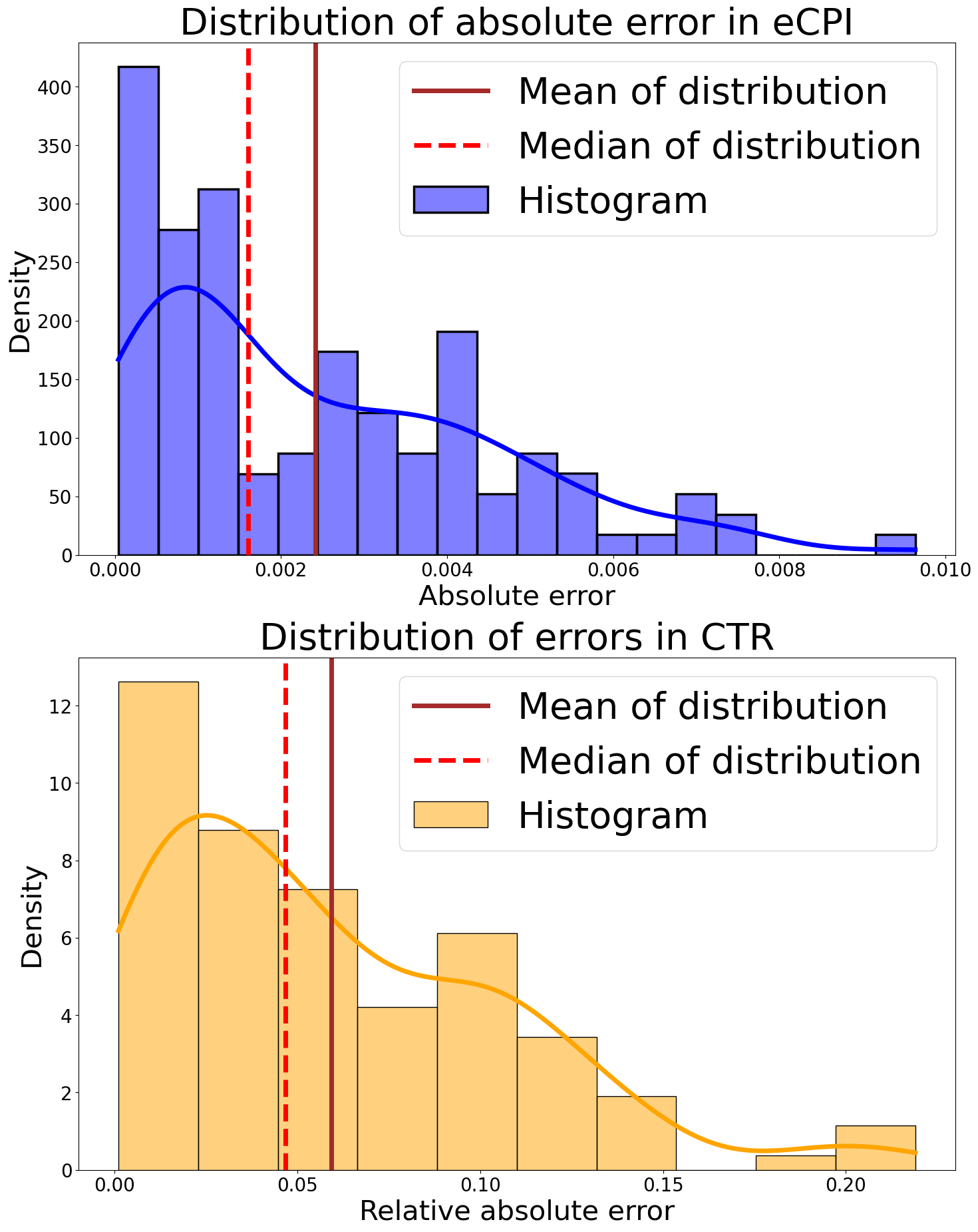}
    \caption{Evaluation of AuctionUCB-PBM on synthetic data, generated as 1(c) combination - fixed $price=1$ and CTR's from real distribution. Upper is distribution of absolute error in eCPI estimation. Lower graphic describes the distribution of absolute relative error in CTR (eCPI)}
    \label{fig::synthetic2}
\end{figure}

For price sampling, the following distributions are employed:

\begin{enumerate}
    \item A degenerate distribution with a price fixed at 1, leading to a straightforward scenario where items are ranked solely by conversions and the increase in Upper Confidence Bound (UCB). 
    \item A uniform distribution ranging from 1 to $K$, which simplifies the task; if conversions are concentrated at a single point, each advertisement's eCPI becomes easily identifiable, allowing algorithms to determine the true order and maximize revenue. 
    \item A binomial distribution with $n=10$ and $p=0.5$, offering an intriguing case due to its symmetry around the expectation, necessitating consideration of the joint distribution of eCPI rather than price alone.
\end{enumerate}

For conversions, the following are utilized: 

\begin{enumerate}[label=(\alph*)]
    \item Uniform CTRs ranging from 0.1 to 0.8, which may pose challenges in learning conversions in certain price distribution combinations. 
    \item Easily distinguishable CTRs, featuring seven items with high conversions (0.8) and others with low conversions (0.1). 
    \item A subsample drawn from real data, representing a particularly compelling case for examination.
\end{enumerate}

Various combinations of the aforementioned distributions were tested, with some findings presented below. Additional experiments are detailed in the $Appendix B$. 

The result of this experiment is shown in Figures \ref{fig::synthetic1}, \ref{fig::synthetic2}. As it can be seen in the Figure \ref{fig::synthetic1} the average $Regret/t$ and instant $Regret$ is decreasing through rounds.  For each graphic standart deviation is added due to average of runs of this experiment. Also, smoothing with $window = 50$ is used for instant regret.

To determine how accurately algorithm was able to estimate the true CTR and eCPI there is presented two distributions:
\begin{itemize}
    \item absolute error in learning eCPI $|eCPI^{true}_i - \hat{eCPI}_i|$ for all agents $i$ with true $eCPI_i^*$ and estimated $\hat{eCPI} = \frac{S_k(t)}{N_k(t)} \cdot price$
    \item relative absolute error of eCPI  $\frac{|eCPI^{true}_i - \hat{eCPI}_i|}{eCPI^{true}}$. Which is equal to relative absolute error in CTR for considering problem.
\end{itemize}


\subsection{Real data} \label{real_data_experiment}

This section evaluates the algorithm using real data from Platform O, which employs a similar sorting method based on the eCPI value. Thus, the proposed algorithm \ref{alg:AuctionUCB-PBM} is applicable within such online systems. Testing was conducted offline on a dataset by logging a portion of auctions, as obtaining thousands of rounds with an identical set of advertisements is not feasible in practice. Instead, advertisements in the upper quantile of opportunities were selected, allowing for rounds with a limited number of ads; rounds with zero participants were omitted, while all others remained valid.
\begin{figure}[h]
    \centering
        \includegraphics[width= 0.5\linewidth]{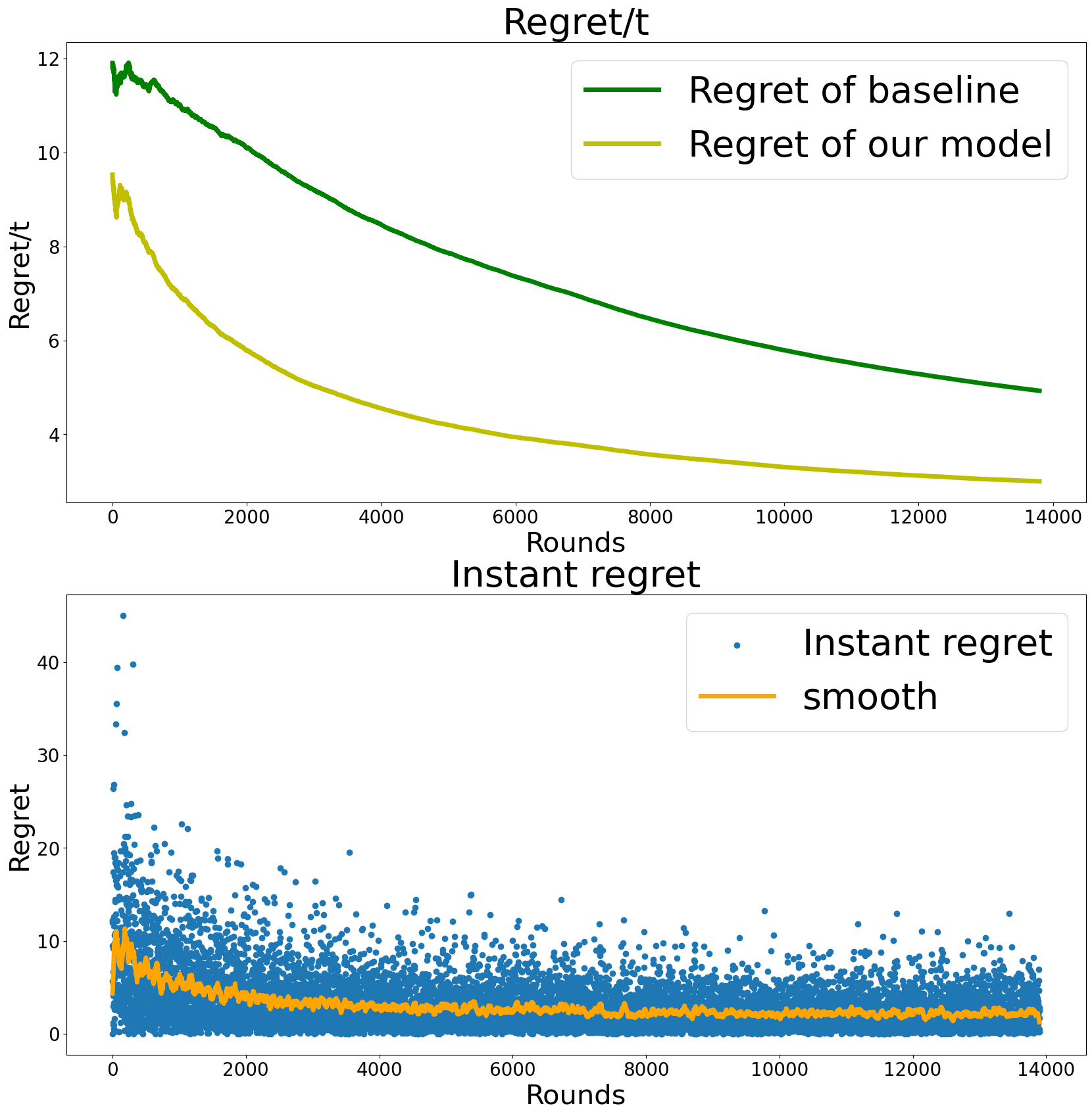}
    \caption{Evaluating AuctionUCB-PBM on real data. Upper graphic describes the average on time cumulative regret comparing baseline strategy with modified $U_k(t, \delta)$ and solution from proposed algorithm\ref{alg:AuctionUCB-PBM}. Lower graphic describes the instant regret and it's smoothed version.}
    \label{fig::real_data1}
\end{figure}
The dataset encompasses information from 21 different regions, although only the 10 regions with the highest number of rounds were considered, as they contain more items with greater opportunities. Each auction is treated as a single step in the algorithm. While updating parameters at each auction can be challenging due to the numerous relevant auctions for each item, this assumption is adopted for the experiment, as it does not impact algorithm convergence and affects cumulative reward. Initially, incorrect estimations for conversions may result in insufficient clicks on high-conversion items and excessive clicks on low-conversion items.

Advertisements from the upper 0.8 quantile were selected from the dataset for each region. Following this initialization, the logged auction results were iterated through to identify participating ads, upon which the AuctionUCB-PBM algorithm \ref{alg:AuctionUCB-PBM} was applied. Ideal users, who click on all ads with a probability equal to their real conversion multiplied by visibility constants, were simulated to obtain auction rewards. Data was then logged, and parameters were updated as described in algorithm \ref{alg:AuctionUCB-PBM}.

To ensure better performance than simpler modifications of the algorithm, results were compared against a baseline algorithm, where 
$U_k (t) = \frac{S_k(t)}{N_k(t)}$
 . Additionally, the goal is to achieve efficiency comparable to the ideal sorting method, from which regret can be calculated.

\begin{figure}[h]
    \centering
        \includegraphics[width=0.5\linewidth]{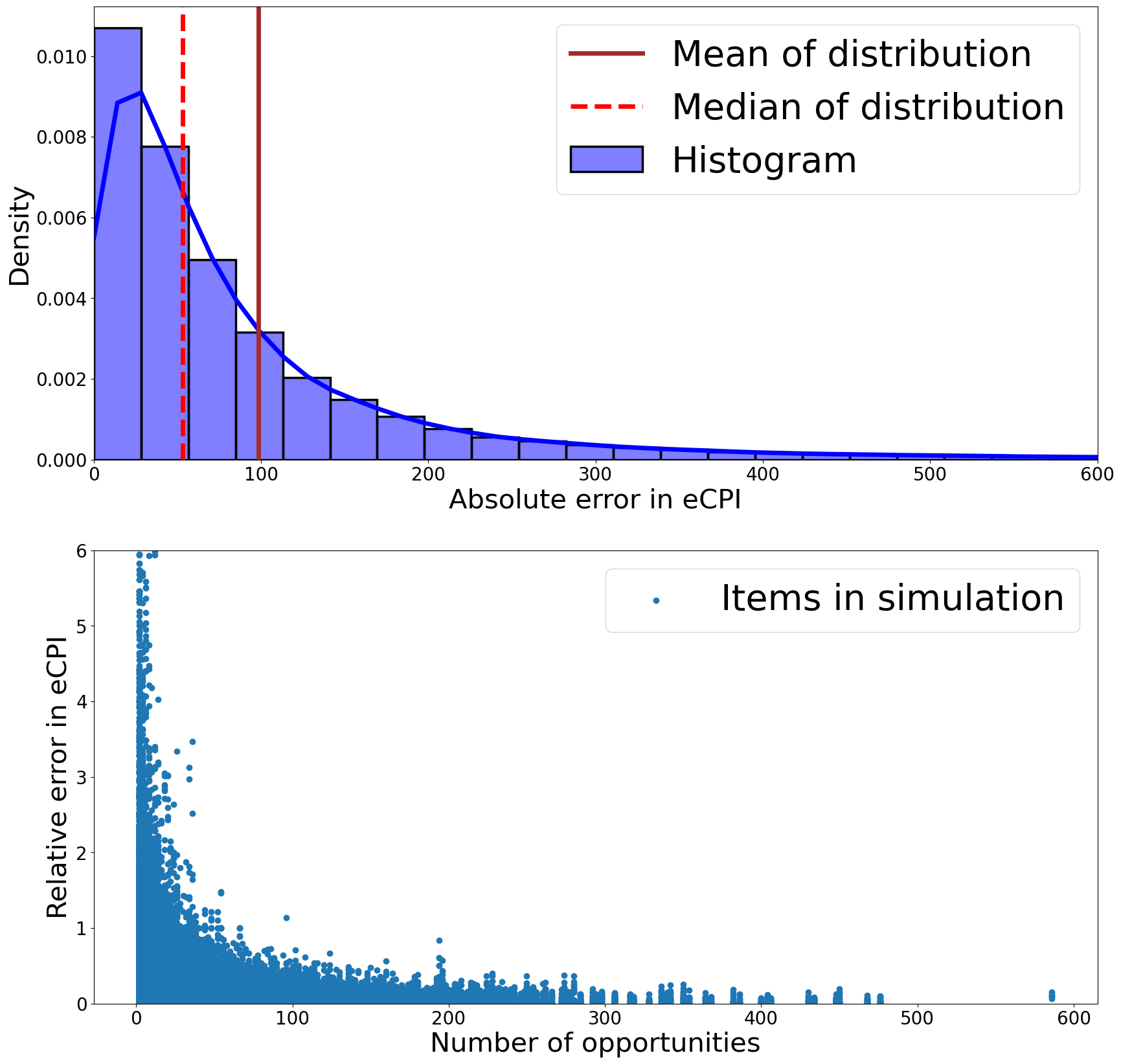}
    \caption{Evaluating AuctionUCB-PBM on real data. The upper picture describes the distribution of absolute error for eCPI's and the lower one desribes the dependence of relative absolute error in CTR (or eCPI's) to number of opportunities.}
    \label{fig::real_data2}
\end{figure}

To align the experiment with real-world conditions, items were divided into two groups: the first group included items with a low number of opportunities, while the second group contained the remaining items. It is assumed that items in the first group possess extensive historical data, enabling reliable estimations of their conversions and eCPIs. Consequently, true estimations are used for these items, and the algorithm is applied solely to items in the second group.

Figure \ref{fig::real_data1} illustrates the expected 
$Regret/t$ and instant regret (the additional regret incurred at each step) for the selected regions. Both metrics exhibit a decreasing trend, thereby supporting the feasibility of implementing the proposed algorithm in a real-world system.  It is observed that only 14,000 to 15,000 steps are deemed the most representative within a daily cycle. As mentioned in the previous section \ref{synthetic_experiment}, the instant regret is smoothed out.  

Figure \ref{fig::real_data2} presents an evaluation of the proposed algorithm \ref{alg:AuctionUCB-PBM} in terms of error distributions. Upper graphic represents distribution of absolute error as it was in previous section. Lower graphic presents dependence of relative absolute error to number of opportunities. So, the bigger opportunities item has the smaller error it has in the of simulation.

\section{"Learning on tails" or how to implement bandits when business ask you to control short-term metric degeneration}

The key reason to use bandit algorithms is to explore items, that didn't get sufficient attention and thus can be underestimated by ML-model. On the other hand exploration can be costly because it can boost positions of both good and bad items. This issue is particularly pressing in online marketing, where users are more likely to abandon a platform if they're not interested in top-recommended products. As a result, customer engagement suffers, and short-term business metrics, such as daily revenue, decline.

To coup with this problem one need to have ability to mix baseline algorithm and bandit algorithm and control this mix. There are two main approaches how to do this in practice. The first is to run A/B-tests on a limited share of traffic. This approach allow to test superiority of the new algorithm in a long run, but do not save from drastic drop in metrics on initial stage before bandits could sort worst items. The other approach is to take solution $x$ from baseline algorithm and solution $y$ from new algorithm and construct mixed solution $z = \lambda x + (1-\lambda) y$ with $\lambda \in (0,1)$. In this case $\lambda$ can be seen as a learning rate and solution $z$ sometimes can be implemented via randomization, i.e. by taking solution $x$ with probability $\lambda$ and solution $y$ with probability $1-\lambda$. This approach can partially solve the problem with potential short-term drop in aggregate revenue metrics, but the main issue with individual user experience is still here since we can't guarantee that mixed solution provide solution comparable to baseline for each individual user.

We present an approach for safely deploying our algorithm in production, leveraging estimates from the existing baseline algorithm. Specifically, we propose use baseline solution on $m$ top slots and to apply bandits only at the tail ($m+1, m+2, \dots)$ of the search results, which reduces the risk of disrupting the performance of the platform for each individual search query.

Since both exploration speed and bandits effect on revenue depends on $m$, we can effectively control implementation risks for bandits on individual query level.  

\textit{Determining the number of top slots.}The number $m$ of top slots can be determined through a parameter $\beta$, which can be interpreted as the level of conservativeness. For instance, let's consider our advertising platform with $L = 30$ slots and visibility coefficients $(\gamma_l)_{l=1,...,L}$ from our experiments \ref{sec:experiments}. By specifying $\beta$ and considering the distribution of visibility coefficients by slots, we can determine the number $m$ of top-performing slots to allocate for our confident items as follows: $m = \min\left(l \in [L]: \frac{\sum_{i = 1}^l{\gamma_i}}{\sum_{i = 1}^L\gamma_i} \geq \beta\right)$. For example, if $\beta = 40\%$, we allocate the top $m = 8$ slots for exploitation (using confident items), as these slots account for approximately $42\%$ of the total visibility $\sum_{l\in [L]}\gamma_l$, leaving a sufficient buffer against exploration of items by UCB-strategy. If $\beta = 30\%$, then we set the value $m = 5$, which cover around $30\%$ of the total visibility.
The choice of threshold $\beta$ significantly impacts how our algorithm is deployed. By setting $\beta$, we can strike a balance between risk tolerance and potential rewards:
\begin{itemize}
\item A lower value of $\beta$ (e.g., $20\%$) would allow us to allocate fewer top-performing slots $(m=3)$ for exploitation, potentially leading to faster speed of items' exploration but also increasing the risk of losses short-term revenue (relative to the baseline).
\item A higher value of $\beta$ (e.g., $80\%$) would require us to allocate more top-performing slots $(m=20)$, potentially leading to slower speed of items' exploration while still maintaining a level of short-term revenue risk tolerance.
\end{itemize}

\textit{CTR estimates for participation in the top-$m$ auction.} Here, we determine which estimates of CTR should be considered acceptable for participation in the auction for top-$m$ positions. 

Let $\theta_k(t)$ denote the estimated Click-Through Rate (CTR) of item $k$ for user in step $t$ in the baseline. Let $\hat\theta_k(t)$ be the empirical CTR score issued by bandits, and $B_k(t)$ be the corresponding confidence interval.

We define a new variable $\theta_k^{\alpha}(t)$ as follows:

$$
\theta_k^{\alpha}(t) = 
 \begin{cases}
   \max\left(\theta_k(t), \hat\theta_k(t)\right) &\text{if $B_k(t) \leq \alpha$},\\
   \theta_k(t) &\text{if $B_k(t) > \alpha$}.
 \end{cases}
$$

Here, $\alpha$ is the acceptable level of confidence interval for items to participate in the auction for the top $m$ positions according to their empirical estimates of CTR.

We rank the items in the first top $m$ positions in descending order by estimating the expected Cost Per Impression (eCPI) as follows:
$$
eCPI^{\alpha}_k(t) = P_k(t) \cdot \theta_k^{\alpha}(t).
$$
For the positions $m+1, m+2, \dots$, we rank the rest of the items based on their UCB estimates of eCPI, which is given by:
$$
\hat{eCPI}_k(t) = P_k(t) \cdot (\hat\theta_k(t) + B_k(t)).
$$

\section{Conclusion}
In this work, we have addressed the critical challenge of cold start in online advertising platforms, where newly introduced and dormant products lack sufficient behavioral data to accurately predict their click-through rates (CTR). By leveraging the multi-armed bandit (MAB) framework and positional-based model (PBM), we have developed an algorithm that effectively allocates ad slots for advertisers in cold-start environments, addressing the challenges of pay-per-click advertising platforms. Our results demonstrate increases in long-term total revenue for online platforms, while also providing theoretical performance guarantees and novel approaches for maintaining short-term profits through controlled exploration and exploitation strategies. We bridge the gap between theoretical learning guarantees and practical cold-start challenges, offering a straightforward and easily implementable solution for real-world online advertising platforms.

\bibliographystyle{unsrtnat}
\bibliography{sample-base} 

\appendix
\section{Proof of $Theorem~1$}
In our regression analysis, we primarily follow the work of \cite{lagree2016multiple} and incorporate additional calculations from \cite{combes2015combinatorial, kveton2015tight}.

We denote by $B_{t,k} = \sqrt{\frac{\delta \ln t}{N_k(t)}}$ the UCB-exploration bonus and by $B_{t,k}^+ = \sqrt{\frac{\delta \ln T}{N_k(t)}}$ an upper bound of this bonus.
\\We define the event $E_t = \{\exists k \in a(t) : \left|\hat \theta_k(t) - \theta_k\right| \geq B_{t,k}\}$. The regret can be decomposed into $$R(T) = \sum_{t=1}^T \Delta_{a(t)}\mathbb{I}_{E_t} + \Delta_{a(t)}\mathbb{I}_{\bar E_t}$$
Now we bound each term in our regret decomposition.
\\
\\ \textbf{First Term.}
\\The first term in our regret decomposition $\sum_{t=1}^T \Delta_{a(t)}\mathbb{I}_{E_t}$. It is small because all of our confidence intervals hold with high probability. In particular, using Chernoff-Hoeffding Bound: $$\mathbb{P}\left( \left|\hat \theta_k(t) - \theta_k\right| \geq B_{t,k}\right)\leq 2\cdot t^{-2\delta}$$ and therefore:
\\
$\mathbb{E}\left[\sum_{t=1}\mathbb{I}_{E_t}\right] \leq \sum_{k\in K}\sum_{t=1}^T\int_{0}^{t} \mathbb{P}\left( \left|\hat \theta_k(t) - \theta_k\right| \geq B_{t,k}\right)ds \leq  \\\leq2\sum_{k\in K}\sum_{t=1}^T\int_{0}^{t} t^{-2 \delta}ds = 2\sum_{k\in K}\sum_{t=1}^T t^{1 - 2\delta} \leq[\delta \geq 1.5]\leq
\\\leq 2\sum_{k\in K}\sum_{t=1}^T t^{-2} \leq \frac{\pi^2}{3}K$. Since $\Delta_{a(t)}\leq \sum_{k=1}^K \gamma_k\cdot P_k:= \Delta_1$ for any $a(t)$, $\sum_{t=1}^T \Delta_{a(t)}\mathbb{I}_{E_t}\leq \frac{\pi^2}{3}K \Delta_1$. So, the first term can be bounded from above in expectation by a constant $C_0(\delta)$ that does not depend on $T$.
\\
\\ \textbf{Second Term.}\\
For each round $t\geq 1$, we define the set of arms
\\$S_t = \{1\leq k \leq K : N_{a_l(t)}\leq \frac{16 \delta P_{max}\left(\sum_{k=1}^K\gamma_k \right)^2}{\Delta^2_{a(t)}}\ln T\}$ and the related events 
\\ $F_t = \{0 < \Delta_{a(t)}\leq 2\sum_{k=1}^K\gamma_k \cdot P_{a_k(t)}\cdot B^+_{t,a_k(t)}\};$
\\ $G_t = \{\left|S_t\right|\geq l\};$
\\ $H_t = \{\left|S_t\right|< l, \exists k \in a(t), N_k(t) \leq \frac{16 \delta P_{max} \left(\sum_{k=1}^l\gamma_k \right)^2}{\Delta^2_{a(t)}}\ln T\}$, where the constraint on $N_k(t)$ only differs from the first one by its numerator which is smaller than the previous one, leading to an even stronger constraint.
\\
\\
\textbf{Fact 1.} $\sum_{t=1}^T\Delta_{a(t)}\mathbb{I}\{\bar E_t, \Delta_{a(t)}>0\} \leq \sum_{t=1}^T\Delta_{a(t)}\mathbb{I}\{F_t\}$.
\\
\textbf{Proof.} Taking action $a(t)$ means that $$\sum_{k=1}^K \gamma_k\cdot P_{a_k(t)}\cdot U_{a_k(t)}(t) \geq \sum_{k=1}^K \gamma_k \cdot P_k \cdot U_k(t) $$
Under event $\bar E_t$, all UCB's are above the true parameter $\theta_k$ so we have 
$\sum_{k=1}^K\gamma_k\cdot P_{a_k(t)}\cdot(\theta_{a_k}(t)+2B_{t,a_k(t)})\geq$\\
$\geq\sum_{k=1}^K\gamma_k\cdot P_{a_k(t)}\cdot(\hat \theta_{a_k(t)}+B_{t,a_k(t)})\geq$\\ $\geq\sum_{k=1}^K\gamma_k\cdot P_k\cdot(\hat \theta_k(t)+B_{t,k}) \geq$\\
$\geq\sum_{k=1}^K \gamma_k \cdot P_k\cdot\theta_k$.
\\
\\
Rearranging the terms above and using $B_{t,k}\leq B^+_{t,k}$, we obtain 
$$2\sum_{k=1}^K\gamma_k\cdot P_{a_k(t)}\cdot B^+_{t,a_k(t)}\geq 2\sum_{k=1}^K\gamma_k \cdot P_{a_k(t)}\cdot B_{t,a_k(t)}\geq \Delta_{a(t)}.$$ Therefore, the event $F_t$ must happen and: $$\sum_{t=1}^T\Delta_{a(t)}\mathbb{I}\{\bar E_t, \Delta_{a(t)}>0\} \leq \sum_{t=1}^T\Delta_{a(t)}\mathbb{I}\{F_t\}$$
\\\textbf{End Proof of $Fact~1$.}
\\
\\
\textbf{Let us show that $F_t \subset (G_t \cup H_t)$ by showing its contrapositive}: if $F_t$ is true then we cannot have $(\bar G_t \cap \bar H_t)$. Assume both of these events are true. Then, we have:\\
$\Delta_{a(t)}\stackrel{F_t}{\leq} 2\sum_{k=1}^K\gamma_k\cdot P_{a_k(t)}\cdot B^+_{t,a_k(t)}$\\
$= 2\sum_{k=1}^K\gamma_k\cdot P_{a_k(t)}\cdot \sqrt{\frac{\delta \ln T}{N_{a_k(t)}}}$\\
$= 2\sqrt{\delta\ln T}\sum_{k=1}^{K}\frac{\gamma_k\cdot P_{a_k(t)}}{\sqrt{N_{a_k(t)}}}$\\
$= 2\sqrt{\delta\ln T}\left(\sum_{k\in \bar S_t}\frac{\gamma_k\cdot P_{a_k(t)}}{\sqrt{N_{a_k(t)}}} + \sum_{k\in S_t}\frac{\gamma_k\cdot P_{a_k(t)}}{\sqrt{N_{a_k(t)}}}\right)$\\
$\stackrel{\bar G_t\cap \bar H_t}{<}2\sqrt{\delta\ln T}\frac{\Delta_{a(t)}}{4\sqrt{\delta\ln T}}\left(\frac{\sum_{k\in \bar S_t}\gamma_k\cdot P_{a_k(t)}}{P_{max}\sum_{k=1}^K \gamma_k} + \frac{\sum_{k\in S_t}\gamma_k\cdot P_{a_k(t)}}{P_{max}\sum_{k=1}^l\gamma_k}\right)$\\
$\leq \Delta_{a(t)}$
\\
which is a contradiction. 

The end of the proof proceeds exactly as in the end of the proof of Theorem: events $G_t$ and $H_t$ are split into subevents corresponding to rounds where each specific suboptimal arm of the list is in $S_t$ or verifies the condition of $H_t$. We define:
\\
$$G_{k,t} = G_t \cap \{N_k(t)\leq \frac{16 \delta P_{max}\left(\sum_{k=1}^K\gamma_k \right)^2}{\Delta^2_{a(t)}}\ln T\},$$
$$H_{k,t} = H_t \cap \{N_k(t)\leq \frac{16 \delta P_{max}\left(\sum_{k=1}^l\gamma_k \right)^2}{\Delta^2_{a(t)}}\ln T\}.$$
The way we define these subevents allows to write the two following bounds:
$$\sum_{k=1}^K\mathbb{I}\{G_{k,t}\} = \mathbb{I}\{G_t\}\sum_{k=1}^K\mathbb{I}\{k\in S_t\}\geq l \mathbb{I}\{G_t\}$$
so, we have: $$\mathbb{I}\{G_t\}\leq \frac{\sum_{k=1}^K G_{k,t}}{l}$$
$$\mathbb{I}\{H_t\}\leq \sum_{k=1}^K \mathbb{I}\{H_{k,t}\}.$$

We can now bound the regret using these two results:
$$\sum_{t=1}^T \Delta_{a(t)}(\mathbb{I}\{G_t\} + \mathbb{I}\{H_t\}) \leq \sum_{t=1}^T\sum_{k=1}^K \frac{\Delta_{a(t)}}{l}\mathbb{I}\{G_{k,t}\} + \sum_{t=1}^T\sum_{k=1}^K \Delta_{a(t)}\mathbb{I}\{H_{k,t}\}$$
$$= \sum_{t=1}^T\sum_{k=1}^K \frac{\Delta_{a(t)}}{l}\mathbb{I}\{G_{k,t}, a(t)\neq a^*\} + \sum_{t=1}^T\sum_{k=1}^K \Delta_{a(t)}\mathbb{I}\{H_{k,t}, a(t)\neq a^*\}.$$

There is a finite number $K!$ of actions. So we decompose each sum above on the different actions in $\mathcal{A}$ possible:
$$...\leq \sum_{t=1}^T\sum_{k=1}^K\sum_{a\in \mathcal{A}} \frac{\Delta_{a}}{l}\mathbb{I}\{G_{k,t}, a(t) = a\}+ \sum_{t=1}^T\sum_{k=1}^K\sum_{a\in \mathcal{A}} \Delta_{a}\mathbb{I}\{H_{k,t}, a(t) = a\}$$
\\The two sums on the right hand side look alike. For arm $k$ fixed, events $G_{k,t}$ and $H_{k,t}$ imply almost the same condition on $N_k(t)$, only $H_{k,t}$ is stronger because the bounding term is smaller. We now rely on a technical result by \cite{combes2015combinatorial} that allows to bound each sum. The Lemma \cite{combes2015combinatorial} for our case will define as:

\textbf{\cite{combes2015combinatorial} Lemma 2.} Let $k$ be a fixed item and $|A| \geq 1, C > 0$, we have
$$\sum_{t=1}^T\sum_{a\in \mathcal{A}} \mathbb{I}\{N_k(t) \leq \frac{C}{\Delta^2_a}, a(t) = a\} \Delta_a \leq \frac{2C}{\Delta_{min}}$$
where $\Delta_{min} = \min_{\Delta_{a} \neq 0} \Delta_a$ is the smallest gap among all suboptimal action. 
\\
\\So, bounding each sum with the above lemma, we obtain:
\\
$... \leq \sum_{k=1}^K\frac{ 32\delta \ln T P_{max}}{\Delta_{min}} \left(\frac{\left(\sum_{k=1}^K \gamma_k\right)^2}{l} + \left(\sum_{k=1}^l \gamma_k\right)^2 \right)= \frac{ 32 K \delta C(l,\gamma) P_{max}}{\Delta_{min}} ln T,$
\\where $C(l, \gamma) = \min_{l=1,..,K}\frac{\left(\sum_{k=1}^K \gamma_k\right)^2}{l} + \left(\sum_{k=1}^l \gamma_k\right)^2$.
\\
\\
So, defining $\delta = 1.5$, we obtain:
$$R(T) \leq \frac{\pi^2}{3}K\Delta_{max} + \frac{ 64 K C(l,\gamma) P_{max}}{\Delta_{min}} ln T.$$
\hfill

\section{Additional experiments}
\begin{figure}[h]
    \centering
        \includegraphics[width= 0.5\linewidth]{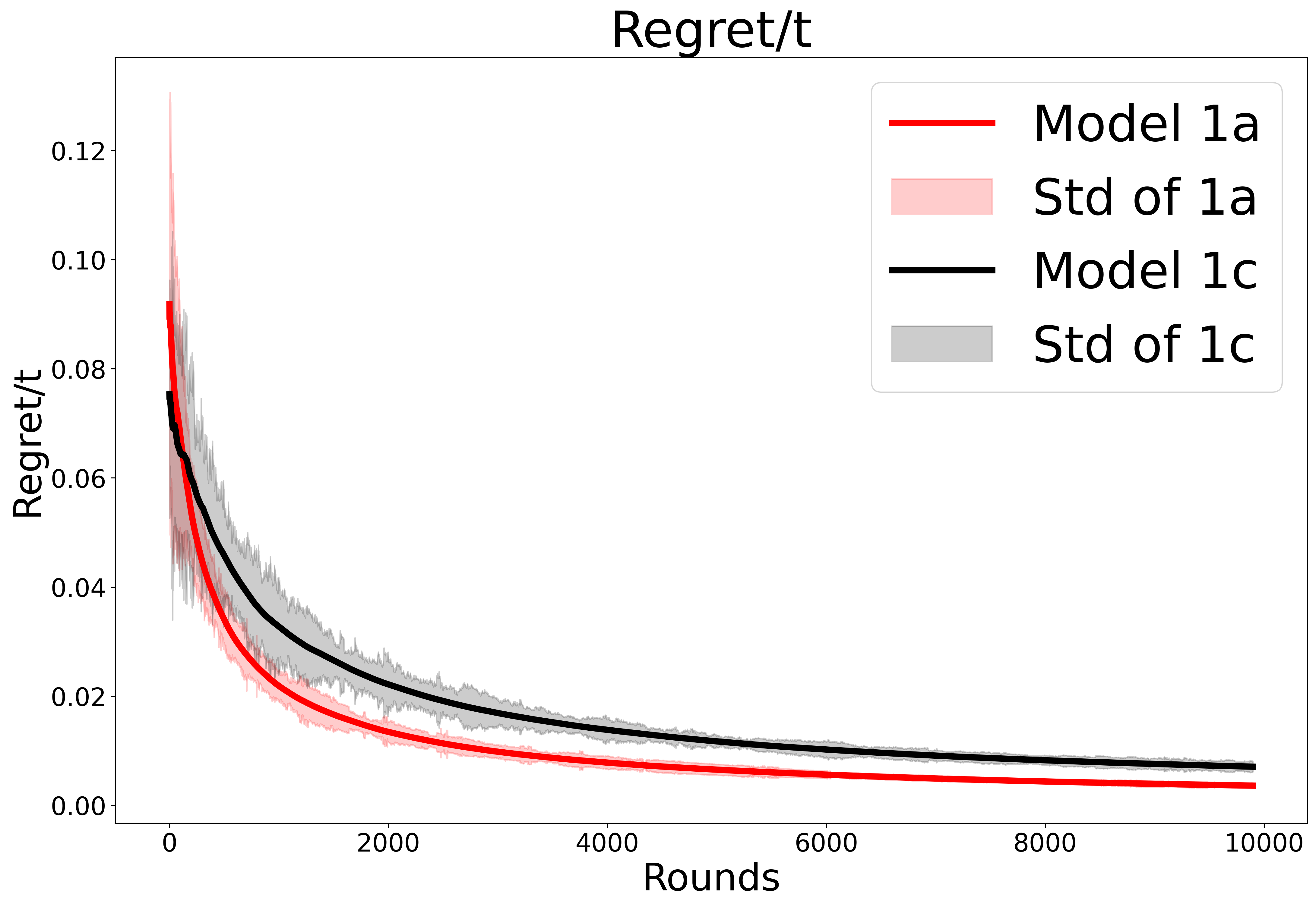}
    \caption{Evaluating AuctionUCB-PBM on synthetic data. Comparing $Regret/t$ on with fixed price - fixed $price=1$}
    \label{fig::synthetic_ap1}
\end{figure}

\begin{figure}[h]
    \centering
        \includegraphics[width= 0.5\linewidth]{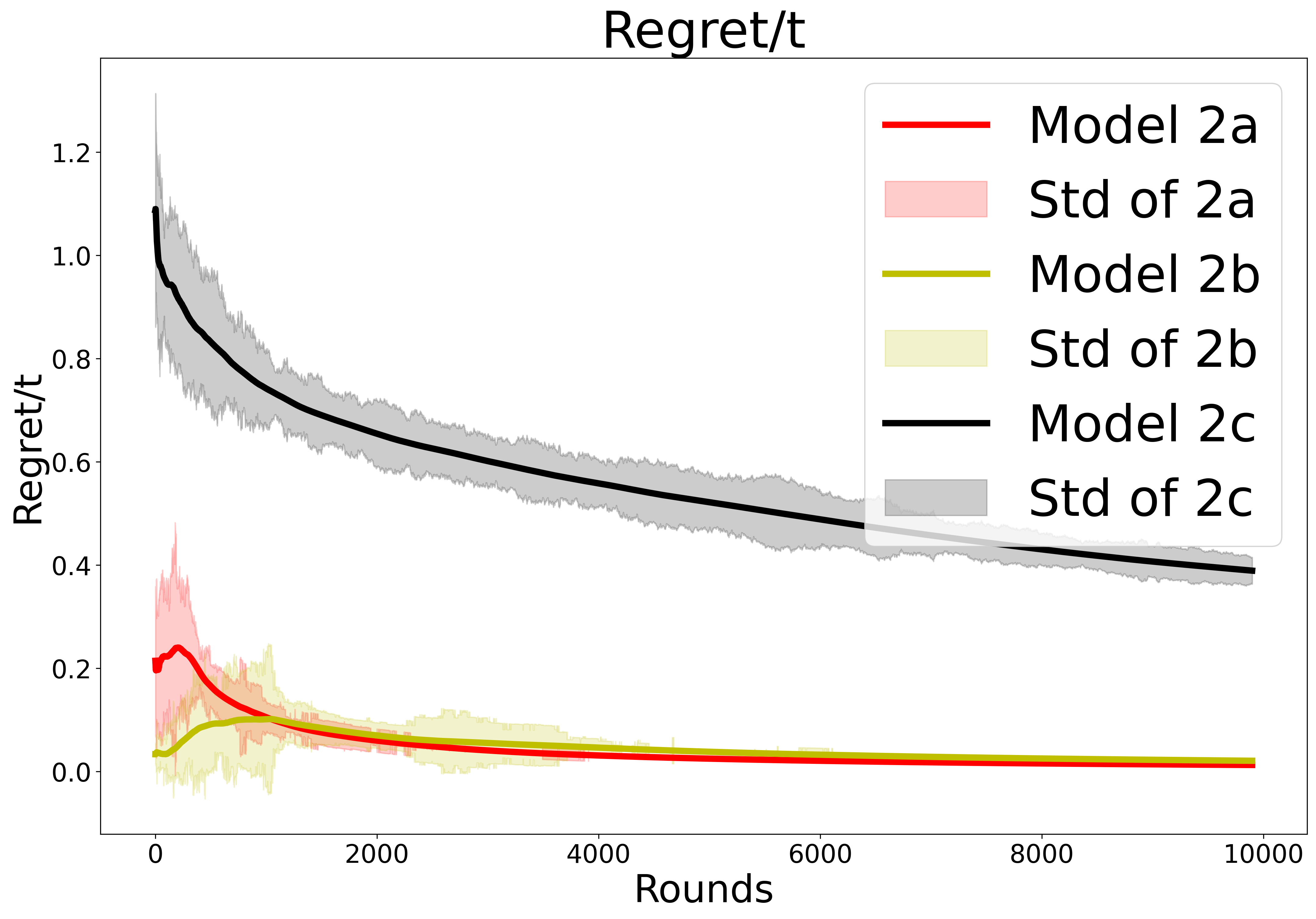}
    \caption{Evaluating AuctionUCB-PBM on synthetic data. Comparing $Regret/t$ on with fixed price - uniform distribution ranging from 1 to $K$}
    \label{fig::synthetic_ap2}
\end{figure}

\begin{figure}[h]
    \centering
        \includegraphics[width= 0.5\linewidth]{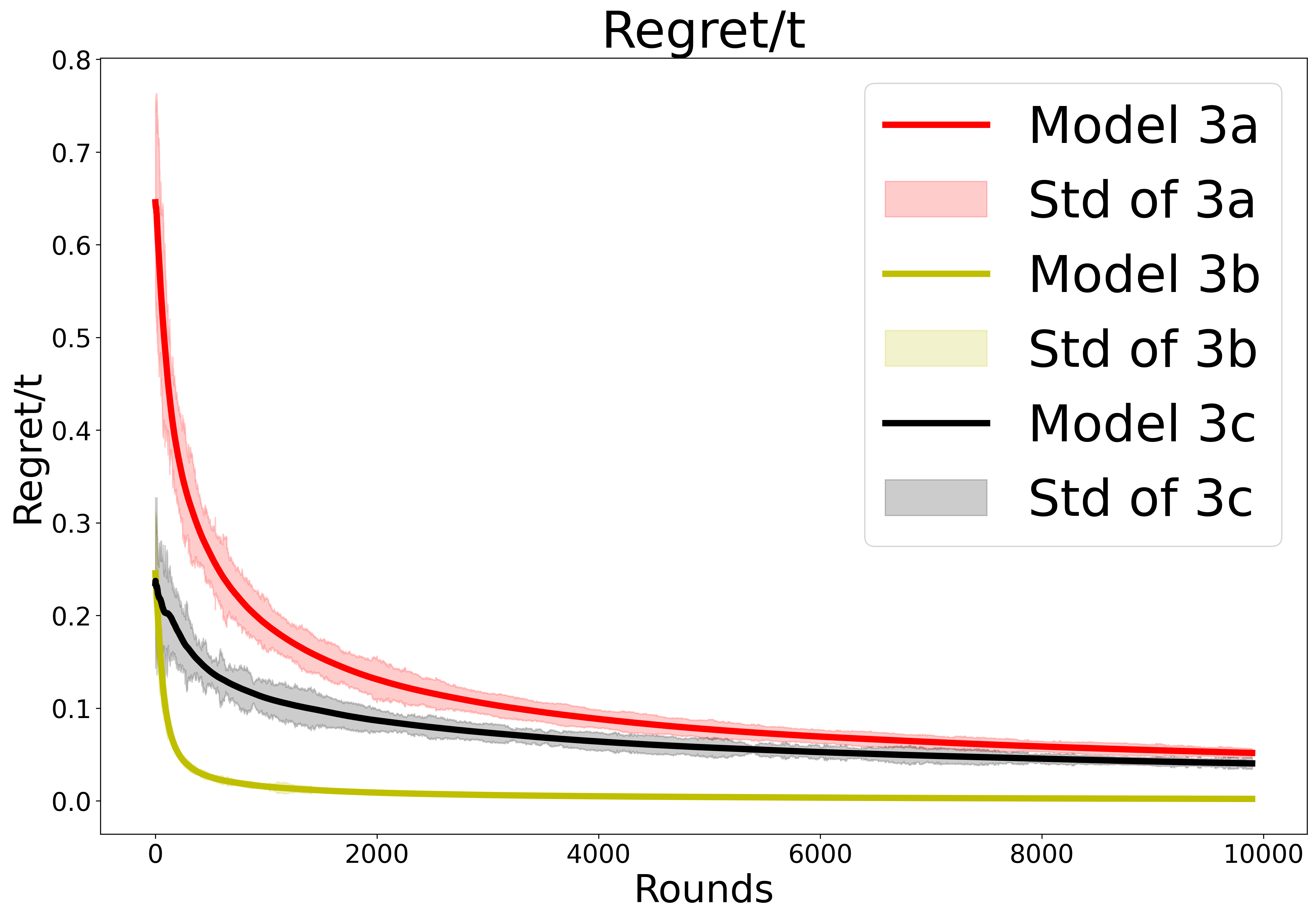}
    \caption{Evaluating AuctionUCB-PBM on synthetic data. Comparing $Regret/t$ on with fixed price - a binomial distribution with $n=10$ and $p=0.5$}
    \label{fig::synthetic_ap3}
\end{figure}

Here is provided some results (see Figures \ref{fig::synthetic_ap1},\ref{fig::synthetic_ap2},\ref{fig::synthetic_ap3}) of evaluation proposed algorithm on different synthetic distributions as it was discussed in section \ref{sec::synthetic}. Also, the case of fixed price and easy distinguishable CTR's is not considered, because proposed algorithm learns true order in first steps and has zero instant regret in most rounds.

The results are quite clear: the setting
$c$ represents the most difficult challenge for the proposed algorithm because it's CTRs are based on real data. Meanwhile, the setting $b$ (only using values $0.1$ and $0.8$) is the easiest, with the exception of the case shown in figure \ref{fig::synthetic_ap2}, which has a uniform price distribution.

\end{document}